\def\year{2019}\relax
\documentclass[letterpaper]{article} 
\usepackage{aaai19}  
\usepackage{times}  
\usepackage{helvet}  
\usepackage{courier}  
\usepackage{url}  
\usepackage{graphicx}  
\usepackage{amsmath}
\usepackage{amssymb}

\usepackage{algorithm}
\usepackage[noend]{algpseudocode}
\usepackage{multirow}
\usepackage{capt-of}
\usepackage{eqnarray,amsmath}

\begin{document}
%
\title{Guided Dropout}

\author{Rohit Keshari, Richa Singh, Mayank Vatsa\\
IIIT-Delhi, India\\
\{rohitk, rsingh, mayank\}@iiitd.ac.in\\
}

\maketitle
 
\begin{abstract}
Dropout is often used in deep neural networks to prevent over-fitting. Conventionally, dropout training invokes \textit{random drop} of nodes from the hidden layers of a Neural Network. It is our hypothesis that a guided selection of nodes for intelligent dropout can lead to better generalization as compared to the traditional dropout. In this research, we propose ``guided dropout'' for training deep neural network which drop nodes by measuring the strength of each node. We also demonstrate that conventional dropout is a specific case of the proposed guided dropout. Experimental evaluation on multiple datasets including MNIST, CIFAR10, CIFAR100, SVHN, and Tiny ImageNet demonstrate the efficacy of the proposed guided dropout. 
\end{abstract}

\section{Introduction}
``Better than a thousand days of diligent study is one day with a great teacher.''
\\[5pt]
\rightline{{\rm ---  Japanese proverb}}

Deep neural network has gained a lot of success in multiple applications. However, due to optimizing millions of parameters, generalization of Deep Neural Networks (DNN) is a challenging task. Multiple regularizations have been proposed in the literature such as $l_1-norm$~\cite{nowlan1992simplifying}, $l_2-norm$~\cite{nowlan1992simplifying}, max-norm~\cite{srivastava2014dropout}, rectifiers~\cite{nair2010rectified}, KL-divergence~\cite{hinton2006fast}, drop-connect~\cite{wan2013regularization}, and dropout~\cite{hinton2012improving},~\cite{srivastava2014dropout} to regulate the learning process of deep neural networks consisting of a large number of parameters. Among all the regularizers, dropout has been widely used for the generalization of DNNs.  

\begin{figure*}[!t]
\centering
  \includegraphics[width=0.8\textwidth]
    {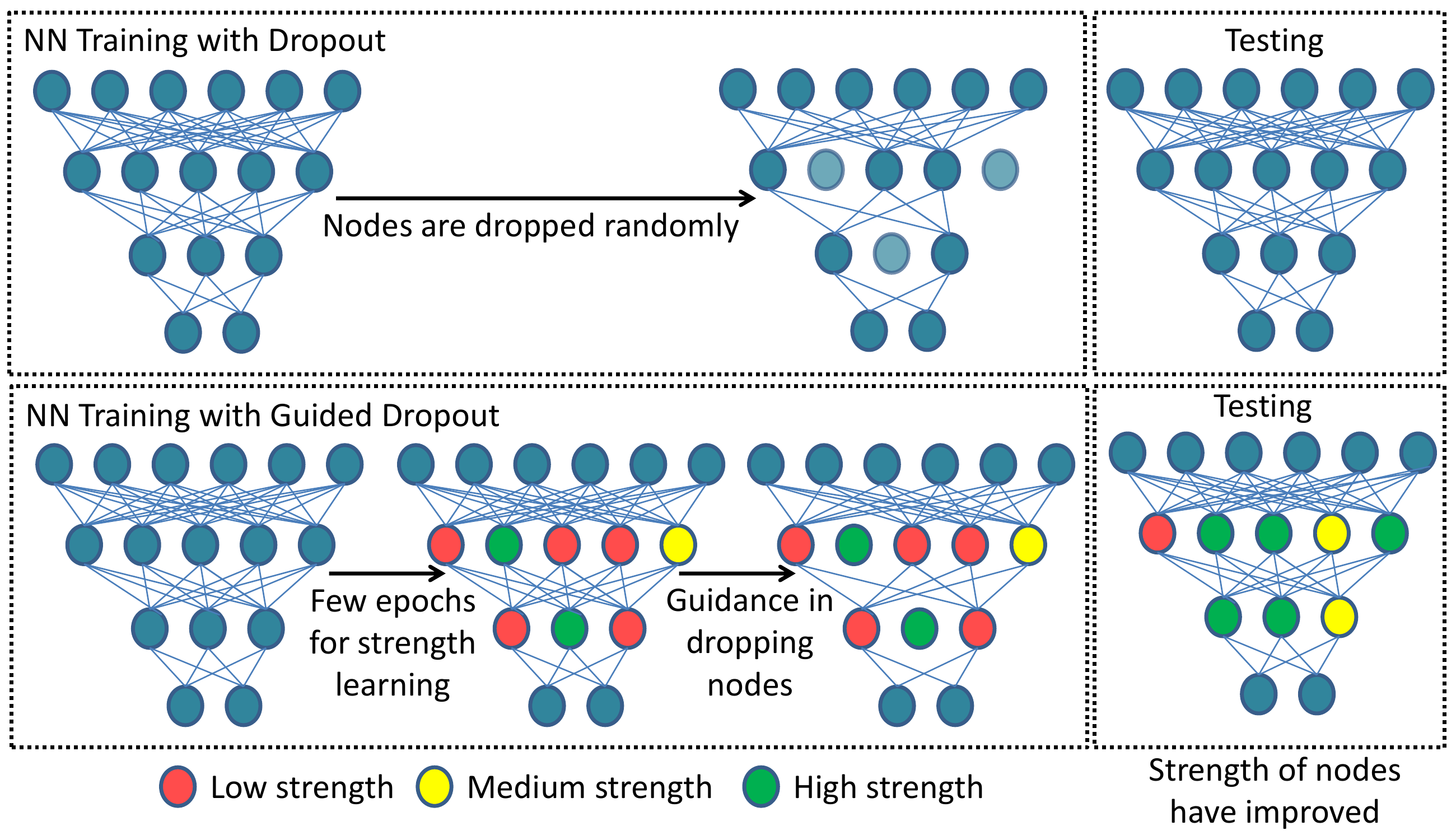}
    \captionof{figure}{Illustrating of the effect of conventional dropout and proposed guided dropout. In neural network training with dropout, nodes are dropped randomly from the hidden layers. However, in the proposed guided dropout, nodes are dropped based on their strength. (Best viewed in color).}	
 	\label{fig:motivation_1}	
 \end{figure*}


Dropout~\cite{hinton2012improving},~\cite{srivastava2014dropout} improves the generalization of neural networks by preventing co-adaptation of feature detectors. The working of dropout is based on the generation of a mask by utilizing $Bernoulli$ and $Normal$ distributions. At every iteration, it generates a random mask with probability ($1-\theta$) for hidden units of the network. 
~\cite{wang2013fast} have proposed a Gaussian dropout which is a fast approximation of conventional dropout.~\cite{kingma2015variational} have proposed variational dropout to reduce the variance of Stochastic Gradients for Variational Bayesian inference (SGVB). They have shown that variational dropout is a generalization of Gaussian dropout where the dropout rates are learned. 

~\cite{klambauer2017self} have proposed alpha-dropout for Scaled Exponential Linear Unit (SELU) activation function.~\cite{NIPS2013_5032} have proposed ``standout" for a deep belief neural network where, instead of initializing dropout mask using $Bernoulli$ distribution with probability $p$, they have adapted the dropout probability for each layer of the neural network. In addition to the conventional learning methods of dropout,~\cite{gal2016dropout} have utilized the Gaussian process for the deep learning models which allows estimating uncertainty of the function, robustness to over-fitting, and hyper-parameter tuning. They have measured model uncertainty by measuring the first and second moments of their approximate predictive distribution.~\cite{gal2017concrete} have proposed ``Concrete Dropout'' which is a variant of dropout where concrete distribution has been utilized to generate the dropout mask. They have optimized the probability $p$ via path-wise derivative estimator.    

\begin{figure}[!t]
 	\centering
 	\includegraphics[width=0.5\textwidth]{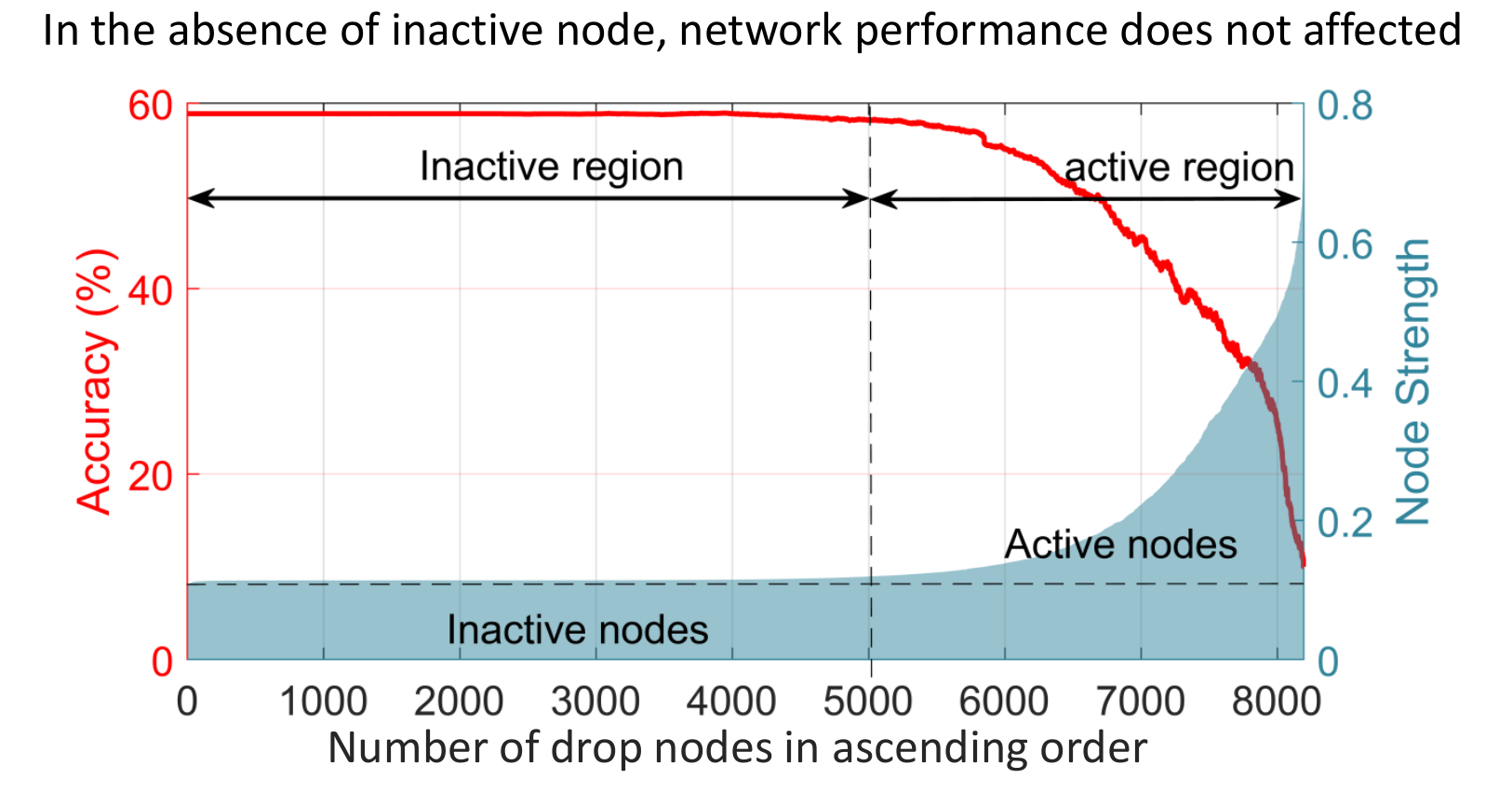}
 	\caption{A Neural Network NN[$8192,3$] is trained with strength parameter on CIFAR10 dataset. The bar graph represents the trained strength of the first layer of the NN. It can be observed that low strength nodes are not contributing in the performance and removing such nodes minimally affect the accuracy. Such nodes are termed as inactive nodes in the inactive region. Similarly, high strength nodes are contributing in the performance and removing such nodes affect the accuracy. Such nodes are termed as active node in the active region. (Best viewed in color).}	
 	\label{fig:active_inactive} 	
 \end{figure}

In the literature, methods related to dropout have been explored in two aspects: 1) sampling dropout mask from different distributions and maintaining mean of the intermediate input while dropping nodes, and 2) adapting dropout probability. However, if prior information related to nodes of a Neural Network (NN) is available, nodes can be dropped selectively in such a way that generalization of NN is improved. Therefore, in this research, we propose ``strength parameter'' to measure the importance of nodes and feature-map for dense NN and CNN, respectively and use it for guiding dropout regularization. Figure~\ref{fig:motivation_1} illustrates the graphical abstract of the paper ``guided dropout''. For understanding the behavior of strength parameter $t$, a three hidden layer neural network with $8192$ nodes is trained with strength parameter using Equation~\ref{eq:nnguided} (details discussed in the next section). After training, the accuracy is evaluated by removing low strength to high strength nodes one by one. The effect on the accuracy can be observed in Figure~\ref{fig:active_inactive}. It shows that removing up to almost $5000$ low strength nodes has minimal affect on the network accuracy. Therefore, such nodes are considered to be \textit{inactive nodes}, lying in the \textit{inactive region}. On removing nodes with high strength, the network accuracy reduces aggressively. Such nodes are considered to be \textit{active nodes} in the \textit{active region}. 

Our hypothesis is that in the absence of high strength nodes during training, low strength nodes can improve their strength and contribute to the performance of NN. To achieve this, while training a NN, we drop the high strength nodes in the active region and learn the network with low strength nodes. This is termed as \textit{Guided Dropout}. As shown in Figure~\ref{fig:motivation_1}, during training the network generalizability is strengthened by ``nurturing'' inactive nodes. Once trained, more nodes are contributing towards making predictions thus improving the accuracy. The key contribution of this paper is: Strength parameter is proposed for deep neural networks which is associated with each node. Using this parameter, a novel guided dropout regularization approach is proposed. To the best of our knowledge, this is the first attempt to remove randomness in the mask generation process of dropout. We have also presented that conventional dropout is a special case of guided dropout, and is observed when the concept of active and inactive regions are not considered. Further, experimental and theoretical justifications are also presented to demonstrate that the proposed guided dropout performance is always equal to or better than the conventional dropout.

\section{Proposed Guided Dropout}

In dropout regularization, some of the nodes from the hidden layers are dropped at every epoch with probability ($1-\theta$). Let $l$ be the $l^{th}$ layer of a network, where the value of $l$ ranges from $0$ to $L$, and $L$ is the number of hidden layers in the network. When the value of $l$ is zero, it represents the input layer, i.e. $a^0=X$. Let the intermediate output of the network be $z^{(l)}$. Mathematically, it can be expressed as:

\begin{equation}
\begin{split}
    z_j^{(l+1)}&=w_{j\times i}^{(l+1)}a_i^l+b_j^{(l+1)} \\
    a_j^{(l+1)}&= f(z_j^{(l+1)})
\end{split}    
\end{equation}

where, $i\in [1,..., N_{in}]$, and $j\in [1,..., N_{out}]$ are index variables for $N_{in}$ and $N_{out}$ at the $(l+1)^{th}$ layer, respectively. $f(.)$ is the $RELU$ activation function. The conventional dropout drops nodes randomly using $Bernoulli$ distribution and is expressed as: $\widetilde{\textbf{a}}^{(l)}= \textbf{r}^{(l)}\odot \textbf{a}^{(l)}$, where $\widetilde{a}$ is the masked output, $a^{(l)}$ is the intermediate output, $\odot$ is the element-wise multiplication, and $r^{(l)}$ is the dropout mask sampled from $Bernoulli$ distribution. While dropping nodes from NN, expected loss $\mathbb{E(.)}$ increases which enforces regularization penalty on NN to achieve better generalization~\cite{mianjy2018implicit}.

\subsection{Introducing strength parameter} 

As shown in Figure~\ref{fig:gen}, in the initial few iterations of training with and without dropout, network performance is almost similar. The effectiveness of dropout can be observed after few iterations of training. Dropping some of the trained nodes may lead to more number of active nodes in the network. Hence, the performance of the network can be improved. Utilizing this observation and the discussion presented with respect to Figure~\ref{fig:active_inactive} (about active/inactive nodes), we hypothesize that a guided dropout can lead to better generalization of a network. The proposed guided dropout utilizes the strength of nodes for generation of the dropout mask. In the proposed formulation, strength is learned by the network itself via Stochastic Gradient Descent (SGD) optimization. Mathematically, it is expressed as:

\begin{equation}
\label{eq:nnguided}
    a_j^{(l+1)}=t_j^{(l+1)}\odot max\left (0,w_{j\times i}^{(l+1)}a_i^l+b_j^{(l+1)}\right )
\end{equation}
where, $\textbf{t}^{(l)}$ is sampled from $uniform$ distribution (assuming all nodes have equal contribution). It can also be used to measure the importance of the feature-map for CNN networks. Therefore, Equation~\ref{eq:nnguided} can be rewritten as:

\begin{equation}
\label{eq:nnguided_cnn}
    \textbf{a}^{(l+1)}=\textbf{t}^{(l+1)}\odot max\left (0,\textbf{a}^l * \textbf{W}^{(l+1)}+b^{(l+1)}\right )
\end{equation}
where, $*$ is a convolution operation, max(0, .) is a RELU operation\footnote{In the case of other activation functions, intermediate feature maps might have negative values. Therefore, $|t|$ (mod of `$t$') can be considered as strength parameter. In this case, `$t$' value approaching to zero represents low strength and node associated with low strength can be considered as an inactive node.}, and $\textbf{a}^l$ is a three-dimensional feature map (for ease of understanding, subscript has been removed).\\[-8pt] 

\textbf{Strength parameter in matrix decomposition:} To understand the behavior of the proposed strength parameter $\textbf{t}$ in a simpler model, let the projection of input $x\in \mathbb{R}^{d_2}$ on $\textbf{W}\in \mathbb{R}^{d_1\times d_2}$ represent label vector $y$ ($\in \mathbb{R}^{d_1}$). Matrix $\textbf{W}$ can be linearly compressed using singular value decomposition (SVD), i.e., $W=Udiag(t)V^T$. Here, top few entries of $diag(t)$ can approximate the matrix $\textbf{W}$~\cite{denton2014exploiting}. This concept can be utilized in the proposed guided dropout.\\[-8pt]

\textbf{Strength parameter in two hidden layers of an NN:}
In an NN environment, let $V\in \mathbb{R}^{d_2\times r}$ and $U\in \mathbb{R}^{d_1\times r}$ be the weight matrices of the first and second hidden layers of NN, respectively. The hypothesis class can be represented as $h_{U,V}(x)=UV^Tx$~\cite{mianjy2018implicit}. In case of \textit{Guided Dropout}, hidden node is parameterized as $h_{U,V,t}(x)=Udiag(t)V^Tx$ which is similar to the SVD decomposition where parameter $\textbf{t}$ is learned via back-propagation. Therefore, $t$ can be considered as a strength of a node which is directly proportional to the contribution of the node in NN performance. 

In this case, the weight update rule for parameters $U$, $V$ and $\textbf{t}$ on the $(s+1)^{th}$ batch can be written as:


\begin{multline*}
U_{s+1}\gets U_s-\\ \eta \left (\frac{1}{\theta}U_s diag(t_s\odot r_s)V_s^Tx_s-y_s\right) x_s^TV_s diag(t_s\odot r_s) 
\end{multline*}
\\[-15pt]
\begin{multline*}
V_{s+1}\gets V_s-\\ \eta x_s \left (\frac{1}{\theta}x_s^T V_s diag(t_s\odot r_s)U_s^T-y_s^T\right)U_s diag(t_s\odot r_s)
\end{multline*}
\\[-15pt]
\begin{multline}
\label{eq:uvt}
diag(t_{s+1})\gets diag(t_{s})-\\ \eta U_s^T \left (\frac{1}{\theta}U_s diag(t_s\odot r_s)V_s^Tx_s-y_s\right)x_s^TV_s
\end{multline}

where, \{($x_s$, $y_s$)\}$^{S-1}_{s=0}$ is the input data, ($1-\theta$) is the dropout rate, $\eta$ is the learning rate, and $r$ is the dropout mask. For the initial few iteration, $r$ is initialized with ones. However, after few iterations of training, $r$ is generated using Equation~\ref{eq:guided_rgen}.

In the proposed algorithm active nodes are dropped in two ways:
\begin{enumerate}
\item Guided Dropout (top-$k$): Select (top-$k$) nodes (using strength parameter) to drop
\item Guided Dropout (DR): Drop Randomly from the active region. 
\end{enumerate}

\textbf{Proposed Guided Dropout (top-k):} While dropping (top-k) nodes based on the strength, the mask for the proposed guided dropout can be represented as:
\begin{equation}
\label{eq:guided_rgen}
    \textbf{r}^l=\textbf{t}^l\leq th, \; where, \; th=\max_{\left \lfloor N\times (1-\theta)\right \rfloor} \textbf{t}^l
\end{equation}
  
  $\max_{\left \lfloor N\times (1-\theta)\right \rfloor}$ is defined as the $k$ large elements of \textbf{t} where, ($1-\theta$) is the percentage ratio of nodes needed to be dropped and $N$ is the total number of nodes. The generated mask $\textbf{r}^l$ is then utilized in equation $\widetilde{\textbf{y}}^{(l)}= \textbf{r}^{(l)}\odot \textbf{y}^{(l)}$  to drop the nodes. Since the number of dropped nodes are dependent on the total number of nodes $N$ and percentage ratio ($1-\theta$), expected loss can be measured by $\mathbb{E}_{b,x}[||y-\frac{1}{\theta}U diag(r)V^Tx||^2]$. If dropout mask $\textbf{r}^l$ drops $top-k$ nodes, the expected loss would be maximum with respect to conventional dropping nodes. Therefore, guided dropout ($top-k$) would impose maximum penalty in NN loss.    
  
\textbf{Proposed Guided Dropout (DR):} The second way of generating guided dropout mask is to select the nodes from the active region, i.e., nodes are Dropped Randomly (DR) from the active region only. Since the number of inactive nodes are large and have a similar strength; therefore, to find the active or inactive region, number of elements in all the bins\footnote{In this case, 100 equally spaced bins are chosen.} have been computed. The maximum number of elements among all the bins is considered as the count of inactive nodes $f_m$. Thus, $f_m$ and $(N-f_m)$ are the number of inactive and active nodes, respectively. Here, ($1-\theta$) is the probability for sampling dropout mask for active region nodes using $Bernoulli$ distribution. Probability with respect to the total number of nodes should be reduced to maintain the mean $\mu$ in the training phase. Therefore, new probability with respect to $N$ is modified as $(1-\frac{f_m}{N}(1-\theta))$. For the proposed guided dropout (DR), when the nodes are dropped randomly from the active region, in Equation~\ref{eq:uvt}, $\frac{1}{\theta}$ will be modified as $\frac{1}{\frac{f_m}{N}(1-\theta)}$. We have carefully mentioned ($1-\theta$) as the percentage ratio for the proposed guided dropout ($top-k$). In case of guided dropout ($top-k$), the generated mask might be fixed until any low strength node can replace the ($top-k$) nodes. On the other hand, for the proposed guided dropout (DR), ($1-\theta$) is the dropout probability.

\subsection{Why Guided Dropout Should Work?}

Dropout improves the generalization of neural networks by preventing co-adaptation of feature detectors. However, it is our assertion that guidance is essential while dropping nodes from the hidden layers. Guidance can be provided based on the regions where nodes are dropped randomly or top few nodes are dropped from the active region. To understand the generalization of the proposed guided dropout, we have utilized Lemma A.1 from~\cite{mianjy2018implicit}. In their proposed lemma: Let $x\in \mathbb{R}^{d_2}$ be distributed according to distribution $D$ with $\mathbb{E}_x[xx^{T}]=\textbf{I}$. Then, for $\mathcal{L}(U, V):=\mathbb{E}_x[||y-UV^Tx||^2]$ and $f(U, V):=\mathbb{E}_{b,x}[||y-\frac{1}{\theta}U diag(r)V^Tx||^2]$, it holds that

\begin{equation}
\label{eq:path_reg}
f(U,V)=\mathcal{L}(U, V)+\lambda \sum_{i=1}^n ||u_i||^2||v_i||^2
\end{equation}
Furthermore, $\mathcal{L}(U, V)=||W-UV^T||^2_F$, where, $diag(r)\in \mathbb{R}^{n\times n}$, $V\in \mathbb{R}^{d_2\times n}$, $U\in \mathbb{R}^{d_1\times n}$, $W\in \mathbb{R}^{d_1\times d_2}$, $\lambda=\frac{1-\theta}{\theta}$. (The proof of this lemma is given in~\cite{mianjy2018implicit}).

\textit{According to the above mentioned lemma, it can be observed that the guided dropout assists NN to have a better generalization}. Let $r$ be sampled from $Bernoulli$ distribution at every iteration to avoid overfitting in conventional dropout. In guided dropout, mask $r'$ is generated based on the strength value $t$. For Equation $\mathcal{L}(U, V)=||W-Udiag(t)V^T||^2_F$, high strength nodes can be chosen to form mask $r'$. Therefore, loss $\mathbb{E}_{b,x}[||y-\frac{1}{\theta}U diag(r')V^Tx||^2]\geq \mathbb{E}_{b,x}[||y-\frac{1}{\theta}U diag(r)V^Tx||^2]$. In this case, penalty would increase while dropping higher strength nodes. The expected loss would be same only if $r=r'$. If $r\neq r'$, the regularization imposed by the proposed guided dropout increases in the training process. Hence, optimizing the loss in the training process helps inactive nodes to improve their strength in the absence of higher strength nodes. 

From Equation~\ref{eq:path_reg}, the path regularization term $\lambda \sum_{i=1}^n ||u_i||^2||v_i||^2$ regularizes the weights $u_i$ and $v_i$ of the inactive node such that the increase in loss due to dropping higher strength nodes can be minimized. Thus, the worst case of generalization provided by the proposed guided dropout should be equal to the generalization provided by the conventional dropout.  

\subsection{Implementation Details}
Experiments are performed on a workstation with two 1080Ti GPUs under PyTorch~\cite{paszke2017automatic} programming platform. The program is distributed on both the GPUs. Number of epoch, learning rate, and batch size are kept as 200, $[10^{-2},...,10^{-5}]$, and 64, respectively for all the experiments. Learning rate is started from $10^{-2}$ and is reduced by a factor of $10$ at every 50 epochs. For conventional dropout, the best performing results are obtained at $0.2$ dropout probability. In the proposed guided dropout, $40$ epochs have been used to train the strength parameter. Once the strength parameter is trained, dropout probabilities for guided dropout (DR) are set to $0.2$, $0.15$, and $0.1$ for $60$, $50$, and $50$ epochs, respectively. However, after strength learning, dropout ratio for guided dropout (top-k) are set to [$0.2$, $0.0$, $0.15$, $0.0$, $0.1$, $0.0$] for [$10$, $40$, $10$, $40$, $10$, $50$] epochs, respectively.  

\section{Experimental Results and Analysis}

The proposed method has been evaluated using three experiments: i) guided dropout in neural network, ii) guided dropout in deep network (ResNet18 and Wide ResNet 28-10), and iii) case study with small sample size problem. The databases used for evaluation are MNIST, SVHN, CIFAR10, CIFAR100, and Tiny ImageNet. 

The proposed guided dropout is compared with state-of-art methods such as Concrete dropout\footnote{\url{https://tinyurl.com/yb5msqrk}}\cite{gal2017concrete}, Adaptive dropout (Standout)\footnote{\url{https://tinyurl.com/y8u4kzyq}}~\cite{NIPS2013_5032}, Variational dropout\footnotemark~\cite{kingma2015variational}, and Gaussian dropout\footnotemark[\value{footnote}]. Alpha-dropout~\cite{klambauer2017self} has also been proposed in literature. However, it is specifically designed for SELU activation function. Therefore, to have a fair comparison, results of the alpha-dropout are not included in Tables. 
\footnotetext{\url{https://tinyurl.com/y8yf6vmo}}

\subsection{Database and Experimental Protocol}
 
\begin{table*}[!t]
\centering
\small
\caption{Test accuracy (\%) on CIFAR10 and CIFAR100~\cite{krizhevsky2009learning} databases using four different architectures of a three layer Neural Network. (Top two accuracies are in bold).}
\label{tb:nn1}
\begin{tabular}{|l|c|c|c|c|c|c|c|c|}
\hline
\multirow{2}{*}{\textbf{Algorithm}}                                        & \multicolumn{4}{c|}{\textbf{CIFAR10}}                             & \multicolumn{4}{c|}{\textbf{CIFAR100}}                            \\ \cline{2-9} 
                                                                           & 1024, 3        & 2048, 3        & 4096, 3        & 8192, 3        & 1024, 3        & 2048, 3        & 4096, 3        & 8192, 3        \\ \hline \hline
Without Dropout                                                            & 58.59          & 59.48          & \textbf{59.72}          & 59.27          & 28.86          & 30.01          & 30.73          & 32.02          \\ \hline
With Dropout                                                               & \textbf{58.77}          & 59.61          & 59.62          & 59.86          & \textbf{31.52}          & \textbf{31.63}          & \textbf{31.37}          & 31.63          \\ \hline
Concrete Dropout                                                          & 57.38          & 57.64          & 57.45          & 55.28          & 28.03          & 29.09          & 28.91          & 31.02          \\ \hline
Adaptive Dropout                                                           & 55.05          & 55.45          & 56.84          & 57.01          & 27.82          & 28.27          & 28.62          & 28.65          \\ \hline
Variational Dropout                                                        & 48.90           & 52.08          & 53.48          & 54.90           & 17.02          & 20.64          & 23.32          & 24.53          \\ \hline
Gaussian Dropout                                                           & 56.12          & 56.52          & 56.94          & 57.34          & 27.24          & 28.34          & 28.87          & 29.81          \\ \hline
Strength only                                                                   & 58.30           & 58.92          & 59.21          & 59.49          & 29.66          & 30.20           & 30.84          & 31.12          \\ \hline \hline
\begin{tabular}[c]{@{}c@{}}Proposed Guided Dropout (top-k)\end{tabular} & 58.75          & \textbf{59.65}          & 59.64          & \textbf{59.92}          & 30.92          & 31.59          & 31.34          & \textbf{32.11}          \\ \hline
\begin{tabular}[c]{@{}c@{}}Proposed Guided Dropout (DR)\end{tabular}    & \textbf{59.84} & \textbf{60.12} & \textbf{60.89} & \textbf{61.32} & \textbf{31.88} & \textbf{32.78} & \textbf{33.01} & \textbf{33.15} \\ \hline
\end{tabular}
\end{table*}

\begin{table*}[!t]
\small
\centering
\caption{Test accuracy (\%) on SVHN~\cite{netzer2011reading} and Tiny ImageNet~\cite{timagenet} databases using four different architectures of a three layer Neural Network (NN). (Top two accuracies are in bold).}
\label{tb:nn2}
\begin{tabular}{|l|c|c|c|c|c|c|c|c|}
\hline
\multirow{2}{*}{\textbf{Algorithm}}                                        & \multicolumn{4}{c|}{\textbf{SVHN}}                                & \multicolumn{4}{c|}{\textbf{TinyImageNet}}                        \\ \cline{2-9} 
                                                                           & 1024, 3        & 2048, 3        & 4096, 3        & 8192, 3        & 1024, 3        & 2048, 3        & 4096, 3        & 8192, 3        \\ \hline \hline
Without Dropout                                                            & \textbf{86.36}          & \textbf{86.72}          & \textbf{86.82}          & 86.84          & 12.42          & 13.74          & 14.64          & 15.21          \\ \hline
With Dropout                                                               & 85.98          & 86.60           & 86.77          & 86.79          & \textbf{16.39}          & 14.28          & 14.69          & 14.44          \\ \hline
Concrete Dropout                                                          & 83.57          & 84.34          & 84.97          & 85.53          & 11.98          & 12.50           & 12.65          & 14.85          \\ \hline
Adaptive Dropout                                                           & 77.67          & 79.68          & 80.89          & 81.96          & 12.41          & 12.98          & 13.75          & 14.17          \\ \hline
Variational Dropout                                                        & 74.28          & 77.91          & 80.22          & 81.52          & 7.95           & 10.08          & 12.91          & 14.69          \\ \hline
Gaussian Dropout                                                           & 72.46          & 78.07          & 80.42          & 80.74          & 13.88          & \textbf{15.67}          & \textbf{15.76}          & 15.94          \\ \hline
Strength only                                                                  & 85.76          & 85.92          & 85.91          & 86.83          & 12.11          & 13.52          & 13.95          & 14.63          \\ \hline \hline
\begin{tabular}[c]{@{}c@{}}Proposed Guided Dropout (top-k)\end{tabular} & 86.12          & 86.57          & 86.78          & \textbf{86.85}          & 15.47          & 15.45          & 15.55          & \textbf{16.01}          \\ \hline
\begin{tabular}[c]{@{}c@{}}Proposed Guided Dropout (DR)\end{tabular}    & \textbf{87.64} & \textbf{87.92} & \textbf{87.95} & \textbf{87.99} & \textbf{17.59} & \textbf{18.84} & \textbf{18.41} & \textbf{17.74} \\ \hline
\end{tabular}
\end{table*}

\begin{figure}[!t]
 	\centering
 	\includegraphics[width=0.48\textwidth]{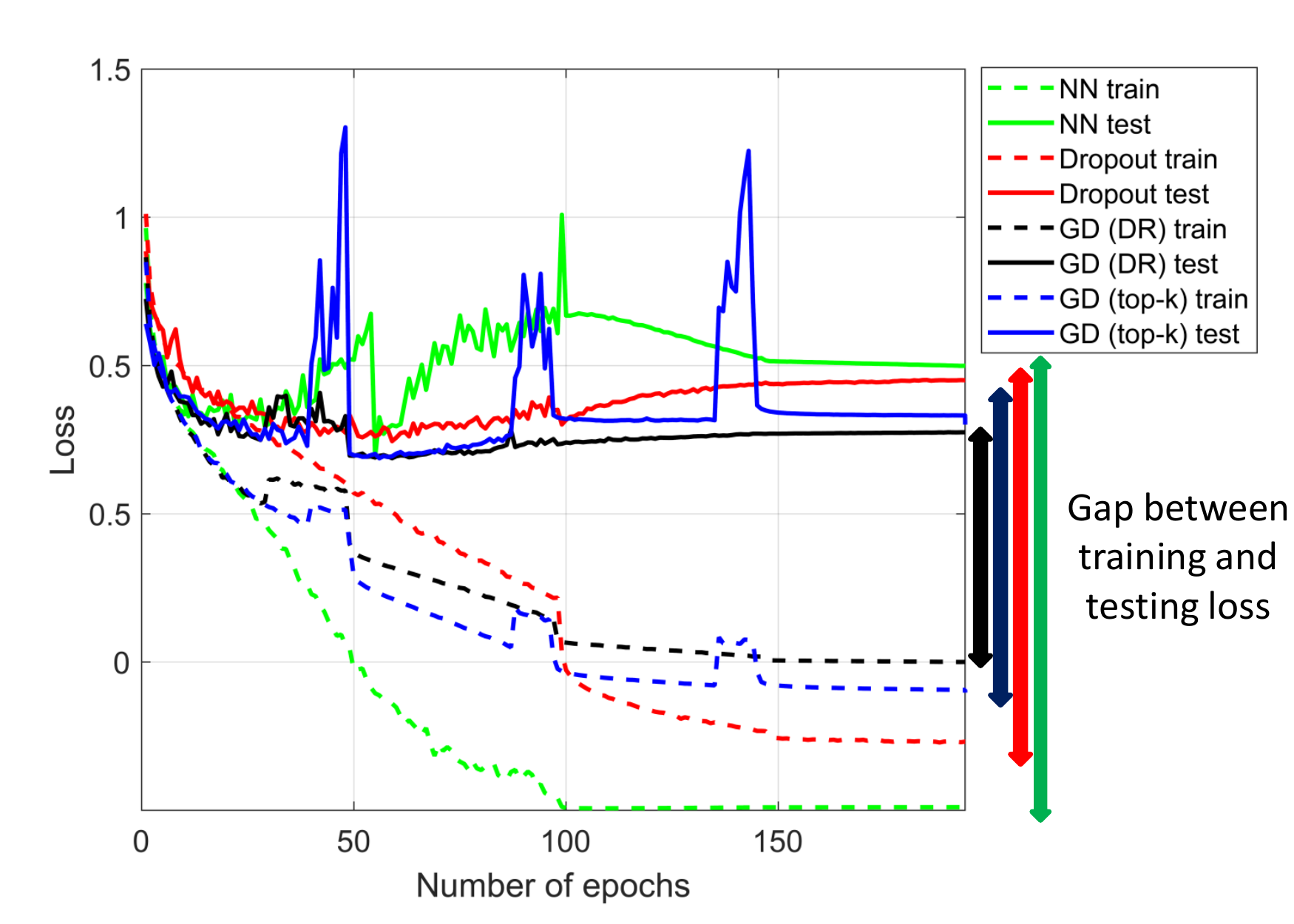}
 	\caption{Illustrating of training and testing losses at every epoch. On the CIFAR10 dataset, the proposed method is compared with the conventional dropout method. It can be observed that the gap between training and testing loss is minimum in proposed guided dropout. (Best viewed in color).}	
 	\label{fig:gen} 	
 \end{figure}

\begin{figure*}[!t]
 	\centering
    \includegraphics[width=1\textwidth]{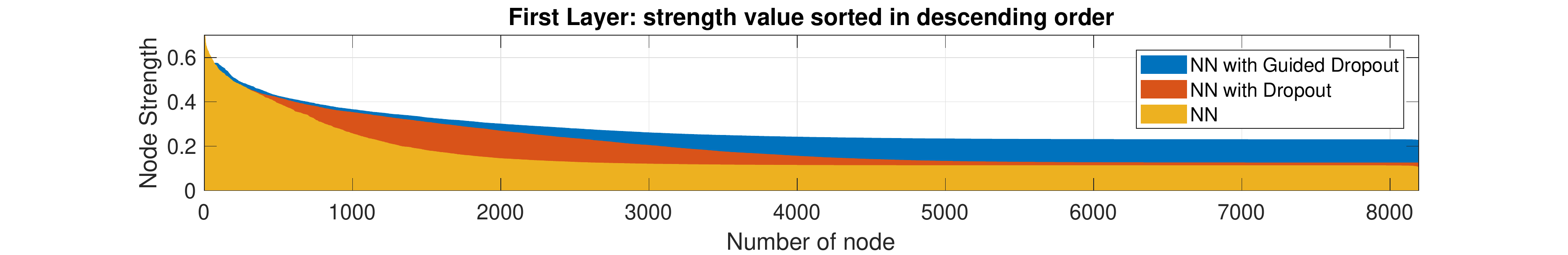}
 	
 	\caption{Illustrating the learned strength values of the first hidden layer nodes for the CIFAR10 database with NN[$8192,3$]. Strength of nodes is improved by utilizing the proposed guided dropout in comparison to with/without conventional dropout. (Best viewed in color).}	
 	\label{fig:str_dist} 	
 \end{figure*}

\textit{Protocol for complete database:} 
Five benchmark databases including MNIST~\cite{lecun1998gradient}, CIFAR10~\cite{krizhevsky2009learning}, CIFAR100~\cite{krizhevsky2009learning}, SVHN~\cite{netzer2011reading}, and Tiny ImageNet~\cite{timagenet} have been used to evaluate the proposed method. MNIST is only used for benchmarking Neural Network with conventional dropout~\cite{srivastava2014dropout}. The MNIST dataset contains $70k$ grayscale images pertaining to 10 classes ($28\times 28$ resolution). The CIFAR10 dataset contains $60k$ color images belonging to $10$ classes ($32\times 32$ resolution). The experiments utilize $50k$ training samples and $10k$ as the test samples. CIFAR100 has a similar protocol with 100 classes. The protocol for CIFAR100 also has $50k$ and $10k$ training-testing split. The SVHN dataset contains $73,257$ training samples and $26,032$ testing samples. Tiny ImageNet dataset is a subset of the ImageNet dataset with 200 classes. It has images with $64\times 64$ resolution with $100k$ and $10k$ samples for training and validation sets, respectively. The test-set label is not publicly available. Therefore, validation-set is treated as test-set for all the experiments on Tiny ImageNet.

\textit{Protocol for small sample size problem:} Recent literature has emphasized the importance of deep learning architecture working effectively with small sample size problems~\cite{keshari2018learning}. Therefore, the effectiveness of the proposed algorithm is tested for small sample size problem as well. The experiments are performed on Tiny ImageNet database with three-fold cross validation. From the entire training set, $200,400...,1k,2k,..,5k$ samples are randomly chosen to train the network and evaluation is performed on the validation set.

\subsection{Evaluation of Guided Dropout in Dense Neural Network (NN) Architecture }

\begin{table}[!t]
\centering
\small
\caption{Test accuracy (\%) on the MNIST~\cite{lecun1998gradient} database using three layer Neural Network (NN). (Top two accuracies are in bold).}
\label{tb:nn-mnist}
\begin{tabular}{|c|c|c|c|c|}
\hline
\multirow{2}{*}{\textbf{Algorithm}}                                        & \multicolumn{4}{c|}{\textbf{Number of nodes, Layers}}                     \\ \cline{2-5} 
                                                                           & \textbf{1024, 3} & \textbf{2048, 3} & \textbf{4096, 3} & \textbf{8192, 3} \\ \hline \hline
Without Dropout                                                            & 98.44            & 98.49            & 98.42            & 98.41            \\ \hline
With Dropout                                                               & 98.45            & \textbf{98.67}            & 98.50             & 98.53            \\ \hline
Concrete Dropout                                                          & \textbf{98.66}            & 98.60             & \textbf{98.62}            & 98.59            \\ \hline
Adaptive Dropout                                                           & 98.31            & 98.33            & 98.34            & 98.40             \\ \hline
Variational Dropout                                                        & 98.47            & 98.55            & 98.58            & 98.52            \\ \hline
Gaussian Dropout                                                           & 98.35            & 98.43            & 98.47            & 98.44            \\ \hline
Strength only                                                                   & 98.42            & 98.51            & 98.40             & 98.46            \\ \hline \hline
\begin{tabular}[c]{@{}c@{}}Proposed Guided \\ Dropout (top-k)\end{tabular} & 98.52            & 98.59            & 98.61            & \textbf{98.68}            \\ \hline
\begin{tabular}[c]{@{}c@{}}Proposed Guided \\ Dropout (DR)\end{tabular}    & \textbf{98.93}   & \textbf{98.82}   & \textbf{98.86}   & \textbf{98.89}   \\ \hline
\end{tabular}
\end{table}

To showcase the generalization of the proposed method, training and testing loss at every epoch is shown in Figure~\ref{fig:gen}. A three layer NN with $8192$ nodes at each layer is trained without dropout, with dropout, and with the two proposed guided dropout algorithms $top-k$, and DR. It can be inferred that the proposed guided dropout approaches help to reduce the gap between the training and testing losses.

The proposed guided dropout is evaluated on three layer Neural Network (NN) with four different architectures as suggested in~\cite{srivastava2014dropout}. Tables~\ref{tb:nn1} to~\ref{tb:nn-mnist} summarize test accuracies (\%) on CIFAR10, CIFAR100, SVHN, Tiny ImageNet, and MNIST databases. It can be observed that the proposed guided dropout (DR) performs better than existing dropout methods. In large parameter setting such as three layer NN with $8192$ nodes, the proposed guided dropout (top-k) algorithm also shows comparable performance. For NN[$1024,3$] architecture, conventional dropout is the second best performing algorithm on CIFAR10, CIFAR100, and Tiny ImageNet databases. 

 \begin{figure}[!t]
 	\centering
    \includegraphics[width=0.5\textwidth]{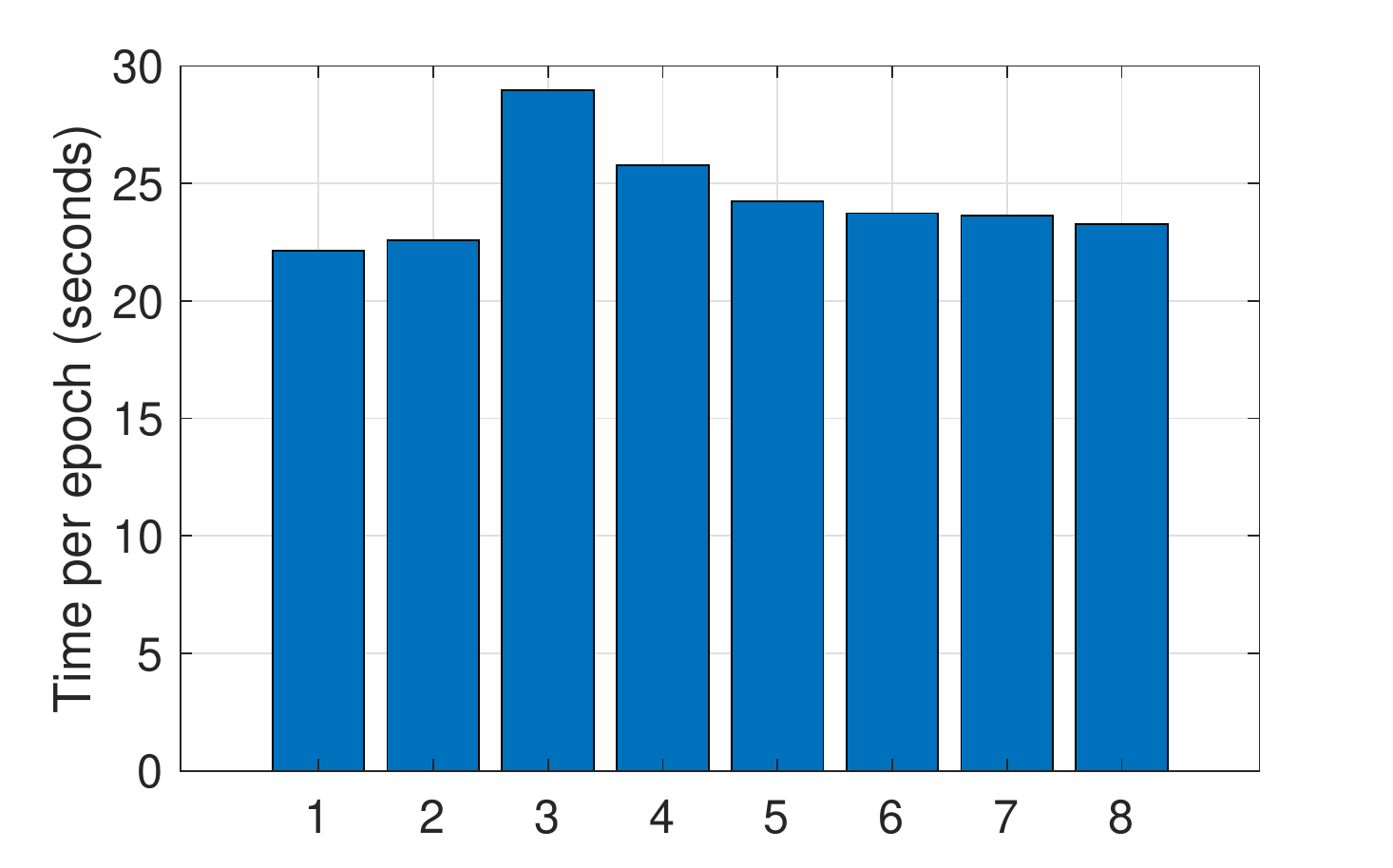}
 	
 	\caption{Illustration of time taken in per epoch. X-axis represents without dropout (1), with dropout (2), concrete dropout (3), adaptive dropout (4), variational dropout (5), Gaussian dropout (6), proposed guided dropout (top-k) (7), and proposed guided dropout (DR) (8) algorithms, respectively.}	
 	\label{fig:time} 	
 \end{figure}
 
We have claimed that the strength parameter is an essential element in NN to measure the importance of nodes. Though the number of training parameters are increased but this overhead is less than $0.2$\% of the total number of parameters of a NN\footnote{For a NN with three hidden layers of $8192$ nodes each, total number of learning parameters is only $8192\times 8192 +8192 \times 8192 =134,217,728$ and overhead of strength parameter is $24,576$.}.

Figure~\ref{fig:str_dist} represents the learned strength value of the first hidden layer of NN[$8196,3$]. It can be observed that the conventional dropout improves the strength of hidden layer nodes. However, the strengths are further improved upon by utilizing the proposed guided dropout. 

For understanding the computational requirements, a NN[$8192,3$] has been trained and time taken without dropout, with dropout, concrete dropout, adaptive dropout, variational dropout, Gaussian dropout, proposed guided dropout (top-k), and proposed guided dropout (DR) and the results are reported for one epoch. Figure~\ref{fig:time} summarizes the time (in seconds) for these variations, which clearly shows that applying the proposed dropout approach does not increase the time requirement.

\subsection{Evaluation of Guided Dropout in Convolutional Neural Network (CNN) Frameworks}

\begin{table*}[!t]
\centering
\caption{Test accuracy (\%) on CIFAR10, CIFAR100~\cite{krizhevsky2009learning} (in Table written as C10, C100), SVHN~\cite{netzer2011reading}, and Tiny ImageNet~\cite{timagenet} databases using CNN architectures of ResNet18 and Wide-ResNet 28-10. (Top two accuracies are in bold).}
\label{tb:cnn}
\begin{tabular}{|c|c|c|c|c|c|c|c|c|}
\hline
\multirow{2}{*}{\textbf{Algorithm}}                                        & \multicolumn{4}{c|}{\textbf{ResNet18}}                                   & \multicolumn{4}{c|}{\textbf{Wide-ResNet 28-10}}                          \\ \cline{2-9} 
                                                                           & \textbf{C10}   & \textbf{C100}  & \textbf{SVHN}  & \textbf{Tiny ImageNet} & \textbf{C10}   & \textbf{C100}  & \textbf{SVHN}  & \textbf{Tiny ImageNet} \\ \hline \hline
Without Dropout                                                            & 93.78          & \textbf{77.01}          & 96.42          & 61.96                 & 96.21          & 81.02          & 96.35          & 63.57                 \\ \hline
With Dropout                                                               & \textbf{94.09}          & 75.44          & \textbf{96.66}          & \textbf{64.13}                 & \textbf{96.27}          & \textbf{82.49}          & \textbf{96.75}          & \textbf{64.38}                 \\ \hline
Concrete Dropout                                                          & 91.33          & 74.74          & 92.63          & 62.95                 & 92.63          & 75.94          & 92.79          & --                    \\ \hline
Adaptive Dropout                                                           & 90.45          & 73.26          & 92.33          & 61.14                 & 79.04          & 52.12          & 90.40           & 62.15                 \\ \hline
Variational Dropout                                                        & 94.01          & 76.23          & 96.12          & 62.75                 & 96.16          & 80.78          & 96.68          & 64.36                 \\ \hline
Gaussian Dropout                                                           & 92.34          & 75.11          & 95.84          & 60.33                 & 95.34          & 79.76          & 96.02          & 63.64                 \\ \hline
Strength only                                                                   & 93.75          & 76.23          & 96.34          & 62.06                 & 95.93          & 80.79          & 96.31          & 64.13                 \\ \hline \hline
\begin{tabular}[c]{@{}c@{}}Proposed Guided \\ Dropout (top-k)\end{tabular} & 94.02          & 76.98          & 96.62          & 64.11                 & 96.22          & 82.31          & 96.42          & 64.32                 \\ \hline
\begin{tabular}[c]{@{}c@{}}Proposed Guided \\ Dropout (DR)\end{tabular}    & \textbf{94.12} & \textbf{77.52} & \textbf{97.18} & \textbf{64.33}         & \textbf{96.89} & \textbf{82.84} & \textbf{97.23} & \textbf{66.02}        \\ \hline
\end{tabular}
\end{table*}

\begin{figure}[!t]
 	\centering
 	\includegraphics[width=0.5\textwidth]{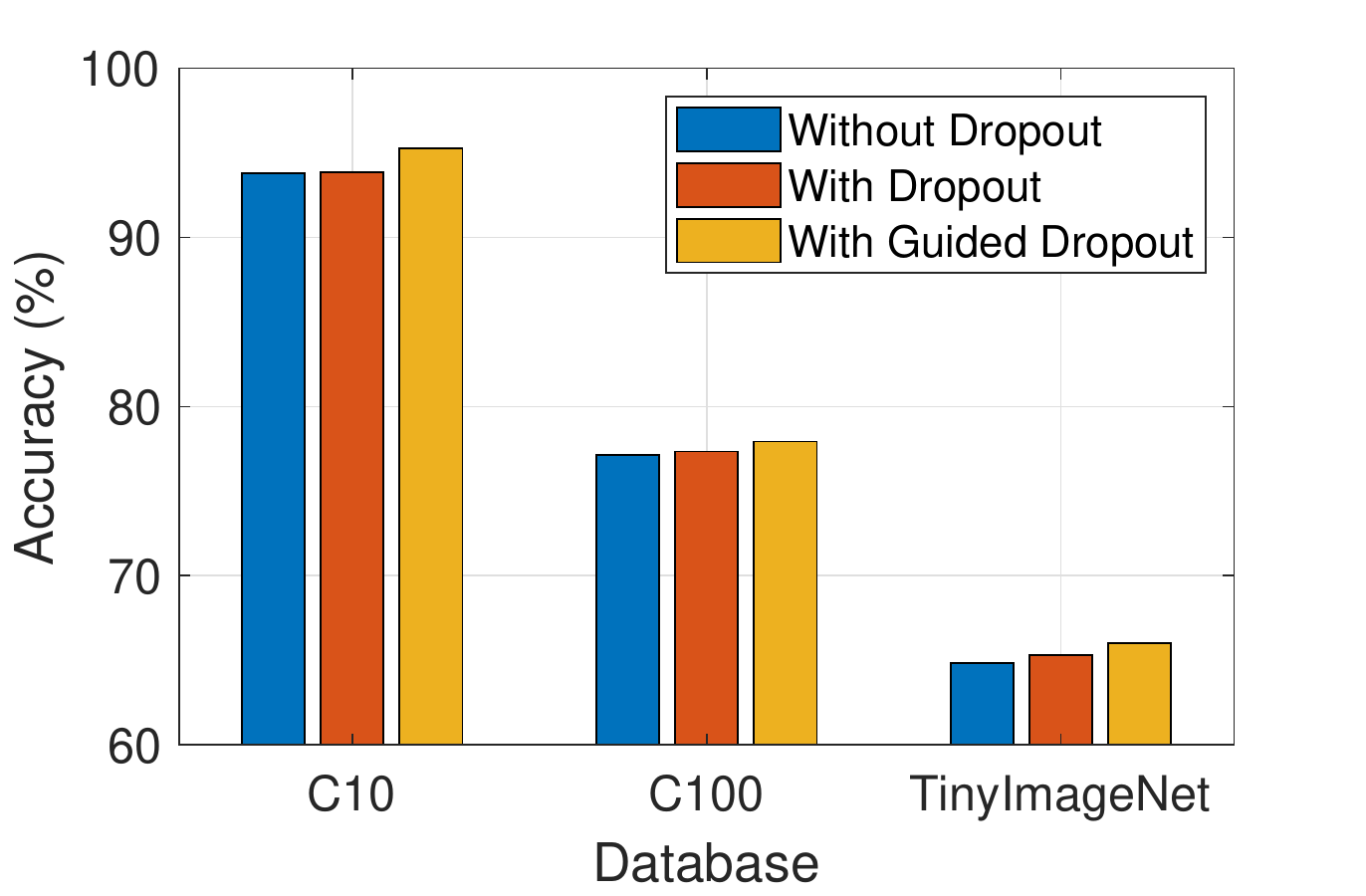}
 	\caption{Classification accuracies on C10, C100, and Tiny ImageNet databases. The performance is measured with ResNet152 CNN architecture without dropout, with traditional dropout, and with the proposed guided dropout. (Best viewed in color).}	
 	\label{fig:res152} 	
 \end{figure}

The proposed guided dropout is also evaluated on CNN architectures of ResNet18 and Wide-ResNet 28-10. On the same protocol, the proposed guided dropout performance is compared with existing state-of-the-art dropout methods. Table~\ref{tb:cnn} summarizes test accuracies of four bench marking databases. It can be observed that on CIFAR10 (C10), dropout is providing second best performance after the proposed algorithm. On CIFAR100 (C100), without dropout is providing second best performance and the proposed guided dropout (DR) is providing the best performance. In case of Wide-ResNet 28-10 which has larger parameter space than ResNet18, conventional dropout consistently performs second best after the proposed guided dropout (DR). It improves the Wide-ResNet 28-10 network performance by $0.62$\%, $0.35$\%, $0.48$\%, and $1.64$\% on C10, C100, SVHN, and Tiny ImageNet databases, respectively. We have also computed the results using ResNet152 CNN architecture on C10, C100, and Tiny ImageNet databases. As shown in Figure~\ref{fig:res152}, even with a deeper CNN architecture, the proposed guided dropout performs better than the conventional dropout method.

\subsection{Small Sample Size Problem}
Avoiding overfitting for small sample size problems is a challenging task. A deep neural network, which has a large number of parameters, can easily overfit on small size data. For measuring the generalization performance of models,~\cite{bousquet2002stability} suggested to measure the generalization error of the model by reducing the size of the training dataset. Therefore, we have performed three fold validation along with varying the size of the training data. 

\begin{figure}[!t]
\small
 	\centering
 	\includegraphics[width=0.49\textwidth]{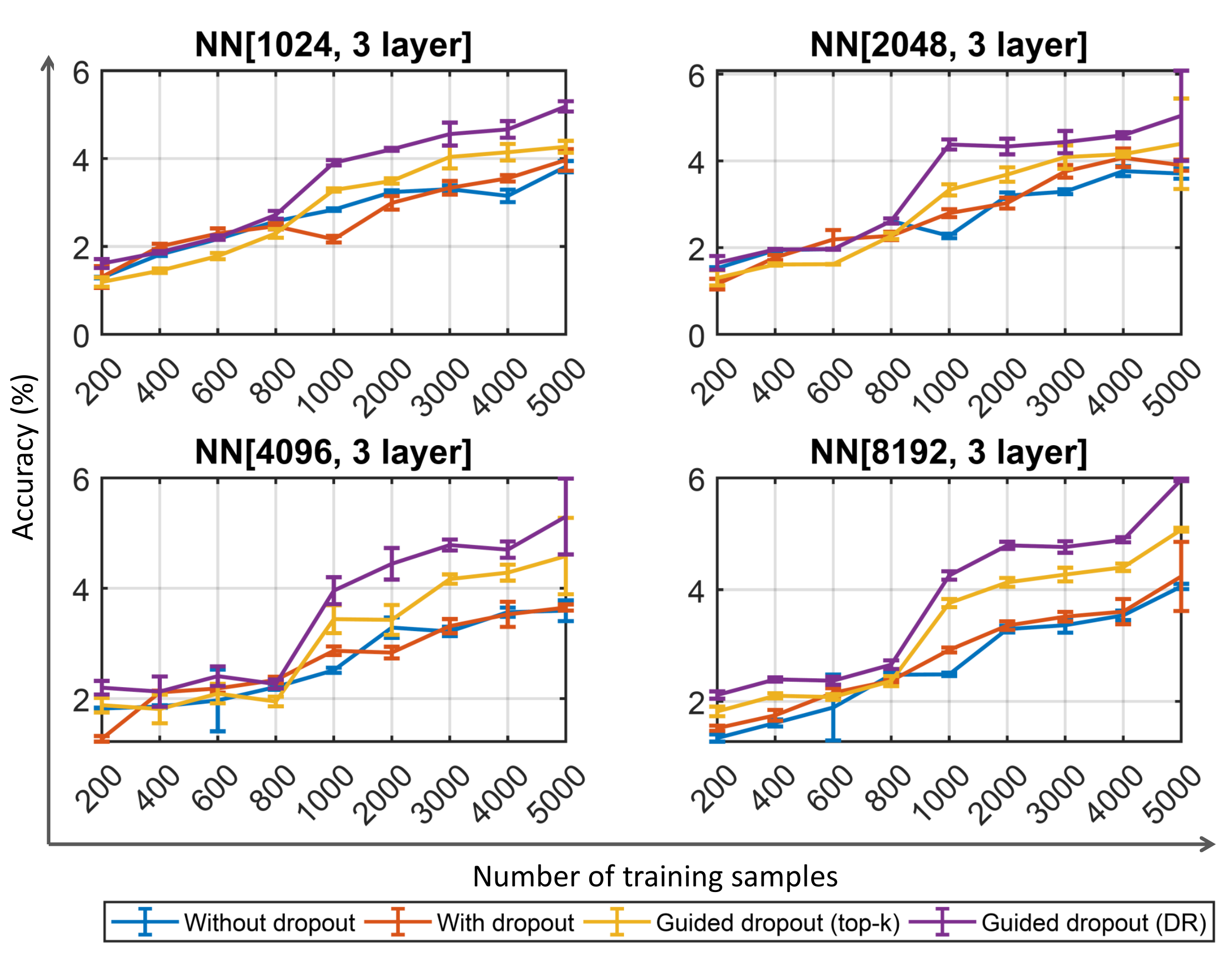}
 	\caption{Results of small sample size experiments: Accuracies on varying training samples of the Tiny ImageNet dataset. The performance has been measured on four different dense NN architectures. (Best viewed in color).}	
 	\label{fig:NNsss} 	
 \end{figure}

\begin{figure}[!t]
 	\centering
 	\includegraphics[width=0.49\textwidth]{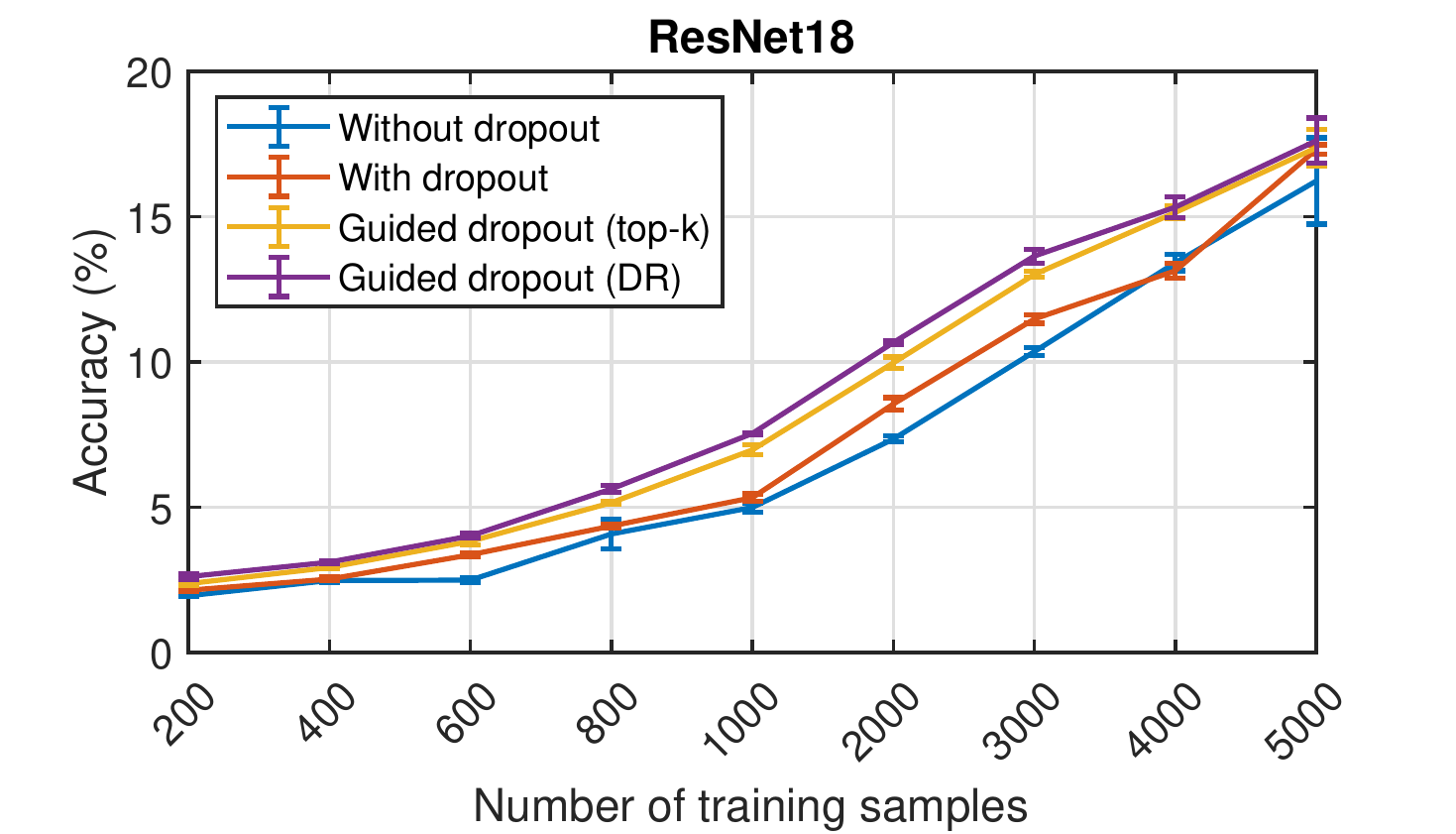}
 	\caption{Classification accuracies obtained with varying the training samples for the Tiny ImageNet dataset. The performance is measured with ResNet18 CNN architecture. (Best viewed in color).}	
 	\label{fig:resNsss} 	
 \end{figure}

The experiments are performed with ResNet-18 and four dense neural network architectures. As shown in Figures~\ref{fig:NNsss} and~\ref{fig:resNsss}, with varying training samples of the Tiny ImageNet database, the proposed guided dropout (DR) yields higher accuracies compared to the conventional dropout. 

\section{Discussion and Conclusion}
Dropout is a widely used regularizer to improve the generalization of neural network. In the dropout based training, a mask is sampled from $Bernoulli$ distribution with ($1-\theta$) probability which is used to randomly drop nodes at every iteration. In this research, we propose a guidance based dropout, termed as guided dropout, which drops active nodes with high strength in each iteration, as to force non-active or low strength nodes to learn discriminative features. During training, in order to minimize the loss, low strength nodes start contributing in the learning process and eventually their strength is improved. The proposed guided dropout has been evaluated using dense neural network architectures and convolutional neural networks. All the experiments utilize benchmark databases and the results showcase the effectiveness of the proposed guided dropout. 

\section{Acknowledgement}
R. Keshari is partially supported by Visvesvaraya Ph.D. fellowship. R. Singh and M. Vatsa are partly supported by the Infosys Center of Artificial Intelligence, IIIT Delhi, India.

\begin{small}
\bibliographystyle{aaai}
\bibliography{dropout}
\end{small}

\end{document}